\documentclass[letterpaper, 10 pt, conference]{ieeeconf}  

\IEEEoverridecommandlockouts                              

\usepackage{graphics} 
\usepackage{epsfig} 
\usepackage{mathptmx} 
\usepackage{times} 
\usepackage{amsmath} 
\usepackage{amssymb}  
 \usepackage{multirow} 
\usepackage{booktabs}
\usepackage{cite} 
\usepackage{booktabs}   
\usepackage{xcolor}      
\usepackage{array}     
\usepackage{url} 
\usepackage{hyperref}
\usepackage{xcolor}
\hypersetup{
    colorlinks=true,
    urlcolor=purple  
}

\usepackage{newtxtext}

\title{\LARGE \bf
WaveFormer: A Lightweight Transformer Model for sEMG-based Gesture Recognition
}

\author{%
  Yanlong~Chen$^{1}$,
  Mattia~Orlandi$^{2}$,
  Pierangelo~Maria~Rapa$^{2}$,
  Simone~Benatti$^{3}$,
  Luca~Benini$^{1}$,
  Yawei~Li$^{1*}$\thanks{$^{*}$Corresponding author: \texttt{li.yawei.ai@gmail.com}}
  \thanks{$^{1}$Integrated Systems Laboratory, ETH Zurich, Z{\"u}rich, Switzerland.}%
  \thanks{$^{2}$DEI, University of Bologna, Bologna, Italy.}%
  \thanks{$^{3}$DIEF, University of Modena and Reggio Emilia, Reggio Emilia, Italy.}%
}

\begin{document}

\maketitle
\thispagestyle{empty}
\pagestyle{empty}

\begin{abstract}
Human-machine interaction, particularly in prosthetic and robotic control, has seen progress with gesture recognition via surface electromyographic (sEMG) signals. However, classifying similar gestures that produce nearly identical muscle signals remains a challenge, often reducing classification accuracy. Traditional deep learning models for sEMG gesture recognition are large and computationally expensive, limiting their deployment on resource-constrained embedded systems. In this work, we propose WaveFormer, a lightweight transformer-based architecture tailored for sEMG gesture recognition. Our model integrates time-domain and frequency-domain features through a novel learnable wavelet transform, enhancing feature extraction. In particular, the WaveletConv module, a multi-level wavelet decomposition layer with depthwise separable convolution, ensures both efficiency and compactness. With just 3.1 million parameters, WaveFormer achieves 95\% classification accuracy on the EPN612 dataset, outperforming larger models. Furthermore, when profiled on a laptop equipped with an Intel CPU, INT8 quantization achieves real-time deployment with a 6.75 ms inference latency. Code and data are available at \url{https://github.com/ForeverBlue816/WaveFormer}.

\end{abstract}

\section{INTRODUCTION}
Surface electromyographic (sEMG) signals are a rich, non‑invasive proxy for muscle activity and support applications ranging from prosthetic control to rehabilitation and gesture‑based human-computer interaction.  Classical sEMG pipelines rely on hand-crafted descriptors that include characteristics of the time domain such as mean absolute value, zero crossings, and waveform length, together with basic spectral indices, which are then fed to shallow classifiers such as $k$ NN, LDA or SVMs~\cite{scheme2011selective}.  Although simple, this scheme has three well‑known limitations.  First, performance is fragile: Even minor electrode shifts or skin impedance changes can dramatically degrade precision because descriptors are highly sensitive to spatial sampling and session‑to‑session drift~\cite{wu2022electrode}.  Second, representing each window with only a few dozen engineered numbers constrains the expressive power of the model, particularly when faced with larger and more diverse datasets that exhibit bursty, non‑stationary patterns and inter‑channel correlations~\cite{samuel2019intelligent}.  Third, competitive results require laborious adjustment of window lengths, feature subsets, and classifier hyper‑parameters, followed by per‑subject recalibration, which impedes broader adoption~\cite{fajardo2021emg,zanghieri2020temponet}.  These shortcomings have spurred interest in end‑to‑end deep architectures that learn robust, multi-resolution representations directly from raw sEMG recordings.  Recent advances in deep learning have shown that these models can outperform classical methods, offering new possibilities for real-time applications~\cite{bakirciouglu2020classification,tsuji2000pattern}.

Motivated by the breakthroughs of large–scale models in language and vision, the time series community has recently proposed several foundation models that learn generic representations from massive sensor corpora. MOMENT is a family of pre-trained masked encoder models trained on \emph{Time‑Series Pile}, achieving strong downstream performance with 385M parameters \cite{goswami2024moment}. OTiS addresses cross‑domain heterogeneity through domain-specific tokenizers and dual masking strategies, achieving state‑of‑the‑art results on 15 benchmarks with 27 billion training samples \cite{turgut2025generalisabletimeseriesunderstanding}. Despite their impressive performance, these models expose two critical gaps for sEMG gesture recognition: (1) parameter counts of tens to hundreds of millions conflict with wearable device constraints, and (2) flat token representations overlook frequency domain cues essential for robust sEMG classification~\cite{wang2014analysis}.

To address these issues, we introduce \textbf{WaveFormer}, a lightweight 3 million parameter Transformer designed specifically for sEMG signals. The model applies a learnable \emph{WaveletConv} front-end for multiscale frequency decomposition, followed by efficient Transformer blocks with rotary positional embedding. By combining frequency‑aware preprocessing with parameter‑efficient attention, WaveFormer achieves the discriminative power of large foundation models while meeting real‑time sEMG system constraints.

Our study makes two primary contributions:
\begin{enumerate} \item We propose WaveFormer, a novel Transformer-based architecture for sEMG gesture recognition, which combines time-domain and frequency-domain features using a learnable wavelet transform. This design enhances the model's ability to extract discriminative features from sEMG signals while maintaining a compact architecture with only 3 million parameters.

\item We demonstrate that WaveFormer achieves state-of-the-art performance in gesture classification across multiple sEMG datasets, outperforming not only much larger foundation models but also lightweight neural architectures and traditional machine learning methods. The model achieves an impressive classification accuracy of 95\% on the EPN612 dataset and 81.93\%  accuracy on the challenging DB6 inter-session protocol, setting new benchmarks in sEMG hand gesture recognition while maintaining a compact 3.10M parameter footprint.
\end{enumerate}
These contributions make WaveFormer a powerful and efficient model for sEMG-based gesture recognition, enabling its deployment in practical applications such as prosthetic control and rehabilitation.

\section{BACKGROUND \& RELATED WORK} 
\subsection{Surface Electromyographic Signals}

Surface EMG (sEMG) records the bioelectric potentials generated by muscle fibers during contraction.  Typical amplitudes range from $10\,\mu\mathrm{V}$ to $1\,\mathrm{mV}$, and most control applications require a bandwidth of up to about $2\,\text{kHz}$; specialized motor unit studies can sample at $\sim\!10\,\text{kHz}$\cite{zanghieri2020temponet}.  Signals are captured non-invasively using gel- or adhesive-based electrodes placed on the skin above the target muscle.  However, measurement quality is highly sensitive to the skin–electrode interface: contact impedance, electrode placement, and adaptation to daily use introduce drift and variability\cite{scheme2011selective}.  Further degradation arises from motion artefacts, cable microphonics, and common‑mode interference when the reference ground is unstable.  Robust sEMG pipelines therefore devote considerable effort to mitigating these noise sources and normalizing recordings across sessions.

\subsection{Wavelet Transform and RoPEAttention}
\begin{figure*}[!t]                
  \centering
  \includegraphics[width=\textwidth]{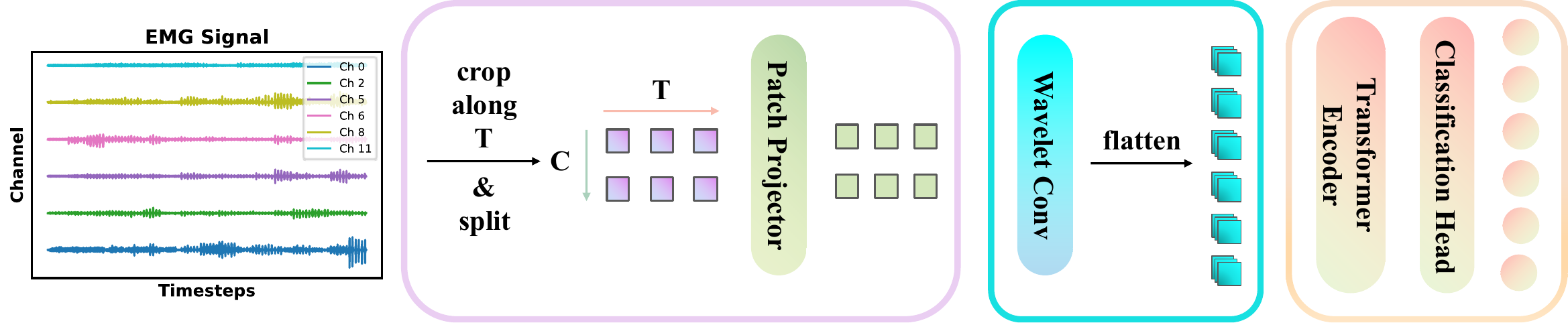} 
  \caption{%
    \textbf{Overview of WaveFormer.} The raw sEMG input 
    $\mathbf{X}\in\mathbb{R}^{C\times T}$ is first segmented into time-domain patches  via a learnable patch projector. These patches are then fed into a \emph{WaveletConv} module that applies multi-level 
    wavelet decompositions and reconstructions to extract rich, multi-scale features. Next, the resulting 
    representations are passed into a Transformer encoder equipped with RoPEAttention to capture global temporal correlations. Finally, a classification head produces likelihood scores for each hand gesture class, which are used to predict the corresponding gesture.
  }
  \label{fig:model}
\end{figure*}

\paragraph{Wavelet Transform}
Wavelet analysis offers a multiscale adaptive time-frequency representation of nonstationary signals~\cite{zhang2019wavelet}. 
Unlike fixed-resolution approaches, the wavelet transform uses dilations and translations of a \emph{mother wavelet} $\psi$ 
to capture both coarse structures and fine transients at different scales. 

For discrete signal processing, the discrete wavelet transform (DWT) decomposes a signal $x[n]$ through a series of filtering and downsampling operations. At each decomposition level $j$, the DWT produces approximation coefficients $A_j$ and detail coefficients $D_j$ via:
\begin{equation}
A_j[k] = \sum_{n} x[n] \cdot h[2k-n], \quad D_j[k] = \sum_{n} x[n] \cdot g[2k-n]
\label{eq:dwt-decomp}
\end{equation}
where $h[n]$ and $g[n]$ are the low-pass and high-pass decomposition filters, respectively. The reconstruction is achieved through:
\begin{equation}
x[n] = \sum_{k} A_j[k] \cdot \tilde{h}[n-2k] + \sum_{k} D_j[k] \cdot \tilde{g}[n-2k]
\label{eq:dwt-reconstruction}
\end{equation}
where $\tilde{h}[n]$ and $\tilde{g}[n]$ are the corresponding reconstruction filters.

In practical sEMG applications, the DWT's multilevel filter–bank implementation splits the raw signal into four two-dimensional subbands at every scale. The low-low (LL) branch, obtained by low-pass filtering across both channel and time axes, stores the slowly varying envelope that reflects general muscle activation trends, and this approximation is recursively decomposed to reveal still coarser behavior. The low–high (LH) band retains rapid temporal transients that are common to all channels, capturing brief motor‑unit bursts; the high–low (HL) band emphasizes spatial differences that change slowly over time, thus disentangling interelectrode crosstalk or electrode placement effects; and the high–high (HH) band isolates short-lived channel-specific high-frequency spikes, including high-frequency noise. Taken together, this coarse‑to‑fine hierarchy preserves global activation patterns while simultaneously exposing localized temporal and spatial details, providing a noise-resistant representation that has proven effective for gesture recognition, onset detection, and artifact suppression under diverse contraction intensities and recording conditions~\cite{finder2024wavelet}.

\paragraph{Rotary Positional Embedding}
Rotary positional embedding (RoPE) provides an alternative to absolute and relative encodings by \emph{rotating} the query ($Q$) and key ($K$) vectors in each self-attention block~\cite{su2024roformer}. Rather than adding explicit positional vectors, RoPE multiplies token embeddings by rotation matrices whose angles depend on token positions.

For a 2D sub-vector at position $m$, RoPE applies the rotation:
\begin{equation}
f_{\{q,k\}}\!\bigl(\mathbf{x}_m, m\bigr) = R(m\theta) \, W_{\{q,k\}} \, \mathbf{x}_m
\label{eq:rope_rotation}
\end{equation}
where $R(m\theta)$ is the rotation matrix:
\begin{equation}
R(m\theta) = \begin{pmatrix}
  \cos(m\theta) & -\sin(m\theta)\\
  \sin(m\theta) & \cos(m\theta)
\end{pmatrix}
\label{eq:rope_matrix}
\end{equation}
and $W_{\{q,k\}}$ represents the learnable linear transforms for queries and keys, with $\theta$ being a fixed constant.

This construction naturally encodes relative positions in attention computations. When computing the dot product $\langle f_q(\mathbf{x}_m,m), f_k(\mathbf{x}_n,n) \rangle$, the relative position information emerges through the rotation difference $R((m-n)\theta)$. For higher-dimensional embeddings, the feature vector is partitioned into 2D pairs, each rotated independently to form a block-diagonal transformation. This approach preserves the standard attention mechanism while incorporating both absolute and relative positional information without explicit position embeddings~\cite{heo2024rotary}.

\section{METHODS}
\subsection{Preprocessing}
\label{sec:data}
Raw sEMG recordings are first filtered and normalized.
We apply a zero-phase bandpass filter to suppress motion artifacts and high-frequency noise (e.g., 20-450\,Hz, 4th order), followed by a powerline notch filter (50/60\,Hz, quality factor $Q{=}30$) to remove mains interference.
Each channel is then standardized via per-channel $z$-score normalization.
The continuous stream is segmented into fixed-length windows using a sliding window (e.g., window length $T{=}1024$ samples with 50\% overlap), and each window is cropped or zero-padded to length $T$ if necessary.
After these steps, each sample is formatted as $\mathbf{X}\in\mathbb{R}^{C\times T}$, where $C$ is the number of electrodes and $T$ the window length.

\subsection{Patch Embedding and 2D Feature Representation}
Consider a preprocessed sEMG segment $\mathbf{X}\in\mathbb{R}^{C\times T}$ with $C$ channels and $T$ time samples.
As illustrated in Fig.~\ref{fig:model}, the signal is first processed by a \emph{patch projector}, a learnable 2D convolution-based embedding layer, that partitions the signal into non-overlapping patches and maps each to a fixed-dimensional latent vector.
Specifically, the patch projector applies a 2D convolution with kernel size $(1,P)$ and stride $(1,P)$, where $P$ is the patch width along the temporal dimension. In our implementation, we use $T=1024$ time samples with patch size $P=64$, resulting in $N=16$ temporal patches.
This operation simultaneously divides the signal into $N{=}T/P$ patches per channel and projects each $P$-dimensional raw patch into an embedding of dimension $D=256$.
The convolution is followed by LayerNorm and GELU activation, yielding a 2D feature map of shape $(D, C, N)$.
This representation preserves both spatial (channel) and temporal relationships, enabling the subsequent wavelet convolution to perform multiscale decomposition along both channel and time dimensions, while maintaining a fine-grained temporal context for efficient transformer processing.

\subsection{WaveletConv}
\begin{figure}[tb]
\centering
\includegraphics[width=1.0\linewidth]{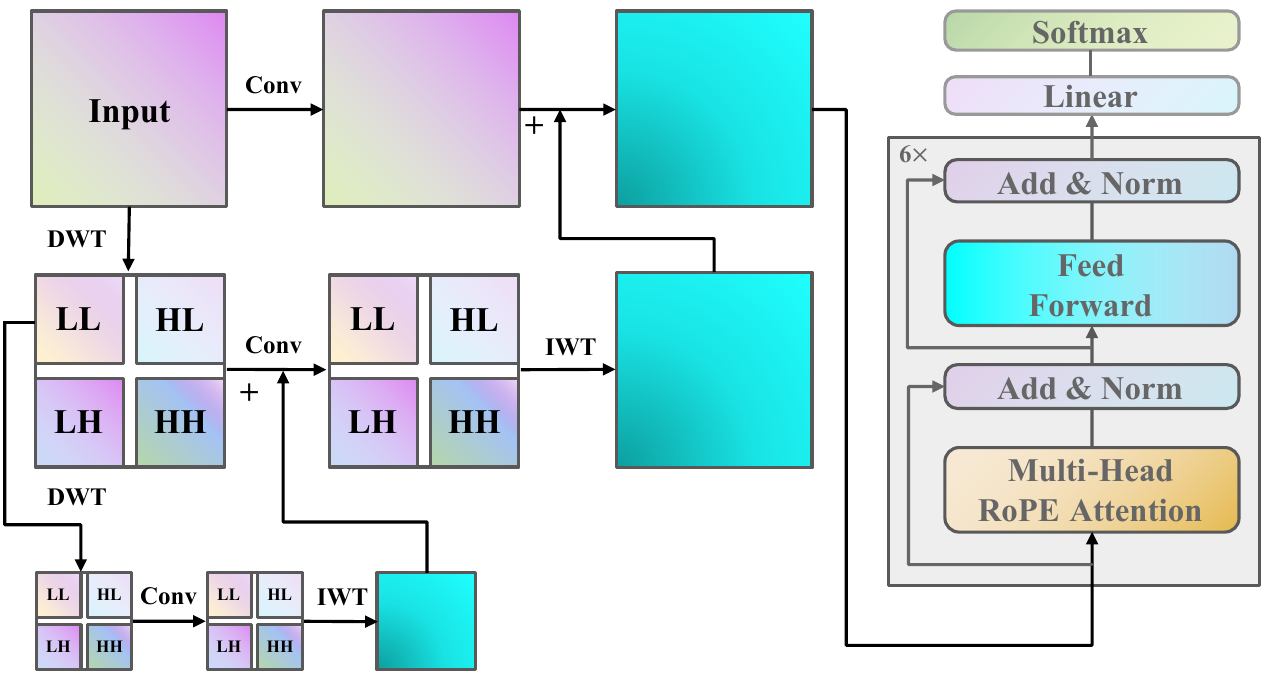}
\caption{%
\textbf{An example WaveletConv pipeline with two levels of wavelet decomposition and reconstruction.} Beginning with an input feature map, a discrete wavelet transform (DWT) splits the signal into four sub-bands (LL, LH, HL, HH) at each level. Each sub-band is processed by learnable depthwise convolutions for frequency-specific feature refinement, with optional dropout applied to high-frequency components. The low-frequency (LL) component undergoes recursive decomposition across multiple levels, while high-frequency components are preserved at each scale. After processing all sub-bands, inverse wavelet transforms (IWT) progressively reconstruct the feature map by combining sub-bands from all decomposition levels. A residual connection ensures preservation of essential baseline information. The resulting wavelet-enhanced features are then flattened and fed into the Transformer encoder with RoPEAttention for gesture classification.
}
\label{fig:wtconv}
\end{figure}

Building upon the 2D feature representation $(D, C, N)$ obtained from patch embedding, WaveFormer applies a learnable wavelet convolution module that exploits the preserved spatial-temporal structure for multiscale decomposition ~\cite{frusque2024robust}.
\paragraph{Overview}
The proposed learnable wavelet transform jointly optimizes decomposition (DWT) and reconstruction (IWT) filters, departing from traditional fixed wavelet bases. As illustrated in Fig.~\ref{fig:wtconv}, the module decomposes the input feature maps into multiple frequency subbands through depth-wise convolutions, then reconstructs the original resolution through transposed depth-wise convolutions. By parameterizing wavelet filters as trainable weights, the network discovers adaptive multiscale decompositions tailored to the characteristics of the sEMG signal.
The architecture performs hierarchical multilevel decomposition, where each level separates low-frequency (LL) and high-frequency (LH, HL, HH) components. Low-frequency components undergo recursive decomposition, creating a pyramid representation that captures both coarse temporal patterns and fine-grained muscle activations. Learnable convolutions process each frequency subband independently, while scaling modules adaptively weight their contributions~\cite{finder2024wavelet}. This end-to-end trainable design enables the model to learn frequency-selective transformations optimized for gesture recognition, achieving superior adaptability compared to fixed wavelet bases~\cite{gao2024efficient}.

\paragraph{Wavelet Filter Learning}
The proposed method initializes 2D decomposition and reconstruction filters from standard wavelet bases (e.g. \texttt{db1})~\cite{vonesch2007generalized}, then optimizes them through backpropagation. During decomposition, depthwise convolutions split each input channel into four frequency subbands (LL, LH, HL, HH), while transposed convolutions perform the inverse operation for reconstruction. This learnable framework enables the filters to evolve beyond their fixed initialization, adapting to the specific spectral characteristics of sEMG signals while maintaining the mathematical properties of wavelet transforms.

\paragraph{Multi-Level Hierarchy}
The architecture employs a three-level wavelet decomposition hierarchy. At each level, only the low-frequency (LL) subband undergoes further decomposition, constructing a pyramid representation from coarse global patterns to fine-grained local variations. High-frequency subbands (LH, HL, HH) are preserved at each scale and optionally regularized through dropout to suppress noise amplification.

Critically, beyond the standard DWT decomposition, we apply additional learnable $(3\times3)$ depthwise convolutions to process the four subbands (LL, LH, HL, HH) at each decomposition level. This dual-stage design first decomposes the signal into frequency subbands via wavelet transforms, then applies a learnable feature refinement within each subband. During training, high-frequency components (LH, HL, HH) undergo selective dropout with probability $p=0.1$ to prevent overfitting to noise patterns while preserving essential discriminative information. This regularization is particularly important for sEMG signals, which often contain high-frequency artifacts of muscle noise and electrode interference~\cite{boyer2023reducing}. The inverse transform progressively reconstructs the signal by combining the processed subbands from all levels of decomposition.

\paragraph{Residual Low-Frequency Path}
A parallel residual path preserves essential low-frequency information that might be attenuated during multilevel decomposition. This path applies a single depthwise convolution to the original input and combines it with the wavelet-reconstructed features:
\begin{equation}
  \mathbf{X}_{\mathrm{final}}
  = \mathrm{Conv}_{\mathrm{base}}(\mathbf{X})
  + \mathbf{X}_{\mathrm{recon}},
  \label{eq:lf-residual}
\end{equation}
where \(\mathrm{Conv}_{\mathrm{base}}(\cdot)\) represents the residual convolution and \(\mathbf{X}_{\mathrm{recon}}\) denotes the wavelet-reconstructed output. This design ensures robust preservation of baseline muscle activity patterns, while the wavelet branch focuses on extracting discriminative temporal dynamics.

\paragraph{Benefits for sEMG}
The learnable wavelet transform provides several advantages for sEMG analysis. First, it adapts to subject-specific and gesture-dependent frequency
distributions without manual basis selection. Second,
multiscale decomposition naturally captures both slow muscle contractions and rapid activation transients characteristic of different gestures. Third, trainable filters can suppress task-irrelevant frequency components while enhancing discriminative patterns. This end-to-end optimization surpasses
fixed wavelet transforms in modeling the non-stationary multi-scale nature of sEMG signals~\cite{zhang2019wavelet,finder2024wavelet}.

\subsection{Transformer-Based Classification via RoPEAttention}
Following the learnable wavelet transform, the refined features $\mathbf{X}_{\mathrm{final}}$ obtained from Eq.~(\ref{eq:lf-residual}) are flattened into a sequence of tokens $\mathbf{Z}$. A learnable class token is prepended to aggregate global temporal-frequency information. The augmented sequence is then processed through a 6-layer Transformer encoder with embedding dimension 256, where each layer employs 8-head Rotary Position Embedding (RoPE) attention.

RoPE directly encodes relative positional information into query and key representations through rotation matrices. Unlike traditional positional embeddings added to input tokens, RoPE applies dimension-wise rotations based on token positions, enabling the model to capture both local muscle activation patterns and long-range temporal dependencies. The Transformer layers follow the standard residual architecture:
\begin{align}
  \mathbf{Z}' &= \mathbf{Z} + \mathrm{RoPEAttn}\bigl(\mathrm{LN}(\mathbf{Z})\bigr), 
  \label{eq:rope-mhsa}\\
  \mathbf{Z}'' &= \mathbf{Z}' + \mathrm{FFN}\bigl(\mathrm{LN}(\mathbf{Z}')\bigr),
  \label{eq:rope-ffn}
\end{align}
where $\mathrm{LN}(\cdot)$ denotes layer normalization and $\mathrm{FFN}(\cdot)$ represents a position-wise feed-forward network with expansion ratio 4, projecting from 256 to 1024 dimensions and back. The RoPE mechanism rotates query-key pairs according to their relative positions, preserving sequential relationships while maintaining computational efficiency.

After processing through all Transformer layers, the class token from the final layer is extracted, normalized, and projected through a linear head to produce gesture logits $\hat{\mathbf{y}}$. The model is trained using cross-entropy loss:
\begin{equation}
  \mathcal{L}_{\mathrm{CE}} = -\sum_{c=1}^{C_{\mathrm{cls}}} y_c\,\ln[\sigma(\hat{y}_c)],
  \label{eq:ce-loss-rope}
\end{equation}
where $y_c$ represents one-hot labels and $\sigma(\cdot)$ is the softmax function. During inference, the predicted gesture corresponds to $\arg\max(\hat{\mathbf{y}})$.

\section{EXPERIMENTAL RESULTS}

\begin{table*}
  \centering
  \caption{Performance comparison (Acc., F1, AUROC) on four sEMG datasets.
  Best results are shown in \textbf{\textcolor{red}{red bold}}, second-best in \textit{italic}.
  Values are mean$\pm$std over 5 runs. * indicates statistical significance over the second-best method ($p<0.05$).}
  \label{tab:emg_performance}
  \scriptsize
  \setlength{\tabcolsep}{4pt}
  \resizebox{\textwidth}{!}{%
  \begin{tabular}{@{}l| c |ccc| ccc| ccc |ccc@{}}
    \toprule
    \multirow{2}{*}{\textbf{Method (year)}} & \multirow{2}{*}{\textbf{\#Params}} &
    \multicolumn{3}{c}{\textbf{EPN612}} & \multicolumn{3}{c}{\textbf{DB5}} &
    \multicolumn{3}{c}{\textbf{DB6}} & \multicolumn{3}{c}{\textbf{UCI EMG}} \\
    \cmidrule(lr){3-5} \cmidrule(lr){6-8} \cmidrule(lr){9-11} \cmidrule(l){12-14}
    & & Acc. & F1 & AUROC & Acc. & F1 & AUROC & Acc. & F1 & AUROC & Acc. & F1 & AUROC \\ 
    \midrule
    Otis (2024)   & 45~M  & 84.82$\pm$1.24 & 85.01$\pm$1.18 & 97.22$\pm$0.31 & 81.79$\pm$1.56 & 60.81$\pm$2.43 & 98.24$\pm$0.22 & 77.96$\pm$1.87 & 77.87$\pm$1.92 & 96.63$\pm$0.45 & 89.14$\pm$0.98 & 89.49$\pm$1.03 & 98.73$\pm$0.18 \\
    Moment (2024) & 385~M & \textit{93.83$\pm$0.76} & \textit{93.75$\pm$0.82} & \textit{99.63$\pm$0.09} & \textit{86.41$\pm$1.13} & \textbf{\textcolor{red}{76.84$\pm$1.95}} & \textit{99.34$\pm$0.12} & \textit{80.36$\pm$1.42} & \textit{80.35$\pm$1.38} & \textit{97.22$\pm$0.28} & \textit{92.79$\pm$0.64} & \textit{91.24$\pm$0.71} & \textit{99.47$\pm$0.11} \\ 
    \midrule[1.2pt]
    WaveFormer (Ours) & 3.10~M & \textbf{\textcolor{red}{95.21$\pm$0.68*}} & \textbf{\textcolor{red}{95.22$\pm$0.71*}} & \textbf{\textcolor{red}{99.70$\pm$0.07*}} &
    \textbf{\textcolor{red}{87.53$\pm$0.92*}} & \textit{74.66$\pm$1.78} & \textbf{\textcolor{red}{99.35$\pm$0.10}} &
    \textbf{\textcolor{red}{81.93$\pm$1.21*}} & \textbf{\textcolor{red}{81.86$\pm$1.25*}} & \textbf{\textcolor{red}{97.69$\pm$0.19*}} &
    \textbf{\textcolor{red}{93.10$\pm$0.58}} & \textbf{\textcolor{red}{93.20$\pm$0.62*}} & \textbf{\textcolor{red}{99.60$\pm$0.08*}} \\
    \bottomrule
  \end{tabular}%
  }
\end{table*}

\subsection{Data and Preprocessing}
To thoroughly evaluate our method, we utilize four publicly available sEMG datasets that offer various conditions and sensor configurations.

EPN‑612 dataset comprises 612 subjects who perform five hand gestures (wave‑in, wave‑out, pinch, open, fist) plus a relaxed state, with each subject providing 50 samples per gesture~\cite{eddy2024big}. Ninapro DB5 dataset includes ten intact subjects who repeat 52 hand movements plus rest, acquired at 200\,Hz using two Thalmic Myo armbands~\cite{pizzolato2017comparison}. Ninapro DB6 focuses on repeatability between sessions, with ten subjects recorded in 2\,kHz using 14 Delsys Trigno electrodes, performing seven grasps over five days~\cite{palermo2017repeatability}. UC Irvine EMG dataset provides raw 200\,Hz sEMG of 36 subjects who perform static hand gestures for 3\,s intervals~\cite{karapinar2021machine}.

All datasets are pre-processed using standard EMG procedures as explained in \ref{sec:data}, including bandpass filtering, notch filtering, $z$ score normalization, and fixed-length window segmentation, resulting in samples of shape $\mathbf{X}\in\mathbb{R}^{C\times T}$. 

The model is trained using the AdamW optimizer with a learning rate of \(4\times10^{-5}\) and a weight decay of \(10^{-4}\). We employ a batch size of 64 to balance GPU memory constraints and gradient stability. Training continues for 30 epochs with a linear warm-up phase of 5 epochs to ensure stable initial convergence. To avoid overfitting, we apply stochastic depth regularization with probability 0.1 and gradient clipping at 1.0 to avoid gradient explosion. Early stopping monitors validation accuracy with patience of 5 epochs and a minimum improvement threshold of 0.01, ensuring optimal model selection without excessive training.

\subsection{Comparison with Foundation Models}
Table~\ref{tab:emg_performance} presents a comprehensive comparison of WaveFormer against leading foundation models (MOMENT and OTiS) across four sEMG datasets. Despite being significantly smaller in parameter count, WaveFormer consistently outperforms both large-scale models. 

WaveFormer demonstrates superior performance across all datasets, achieving 95.21\% precision on EPN612, 87.53\% precision on the challenging 52-class Ninapro DB5, 81.93\% accuracy on the cross-session DB6 evaluation, and 93.10\% accuracy on UCI EMG. These results highlight that our frequency-aware design with learnable wavelet transforms and RoPE-enhanced attention delivers state-of-the-art accuracy while maintaining parameter efficiency ideal for resource-constrained sEMG applications.

\subsection{Comparison with Traditional and Lightweight Models}

Table~\ref{tab:db6_comparison} contrasts WaveFormer with four representative baselines under the demanding DB6 inter-session protocol, in which models are trained on the first five recording sessions and tested on the subsequent five sessions acquired over five days.

Random Forest (25.40\%) is based on hand-crafted features of the time domain, such as the mean absolute value and the length of the waveform extracted from sliding windows~\cite{palermo2017repeatability}. LDA template matching (78.00\%) constructs gesture templates from similar statistical features, applies linear discriminant analysis for dimensionality reduction, and performs classification through nearest-template voting~\cite{li2024high}. TEMPONet (65.20\%) employs a lightweight temporal convolutional network specifically designed for steady-state sEMG segments~\cite{zanghieri2020temponet}. BioFormer (65.73\%) is a Transformer with approximately two million parameters, first pre-trained using an inter-subjects self-supervised protocol on multiple sEMG recordings and subsequently fine-tuned on the NinaPro DB6 dataset~\cite{burrello2022bioformers}.

Against these methods, WaveFormer achieves 81.93\% cross-session accuracy, a substantial 3.93\% improvement over the previous best result, while maintaining a compact 3.10M parameter footprint. This advancement demonstrates WaveFormer's superior ability to learn adaptive frequency-aware representations that remain robust across session-to-session variations in electrode placement, skin impedance, and muscle fatigue, thereby advancing practical sEMG gesture recognition systems.

\begin{table}[tb]
\centering
\caption{Performance comparison on NinaPro DB6 inter-session protocol}
\label{tab:db6_comparison}
\begin{tabular}{lcc}
\hline
\textbf{Method} & \textbf{Parameters} & \textbf{Accuracy (\%)}\\
\hline
Random Forest                & --     & 25.40\\
LDA Template Matching        & --     & 78.00\\
TEMPONet                     & 0.46\,M & 65.20\\
BioFormer                    & 2.00\,M & 65.73\\
\textbf{WaveFormer (Ours)}   & \textbf{3.10\,M} & \textbf{81.93}\\
\hline
\end{tabular}
\end{table}

\subsection{Ablation Study}
\begin{figure}
  \centering
  \includegraphics[width=\columnwidth]{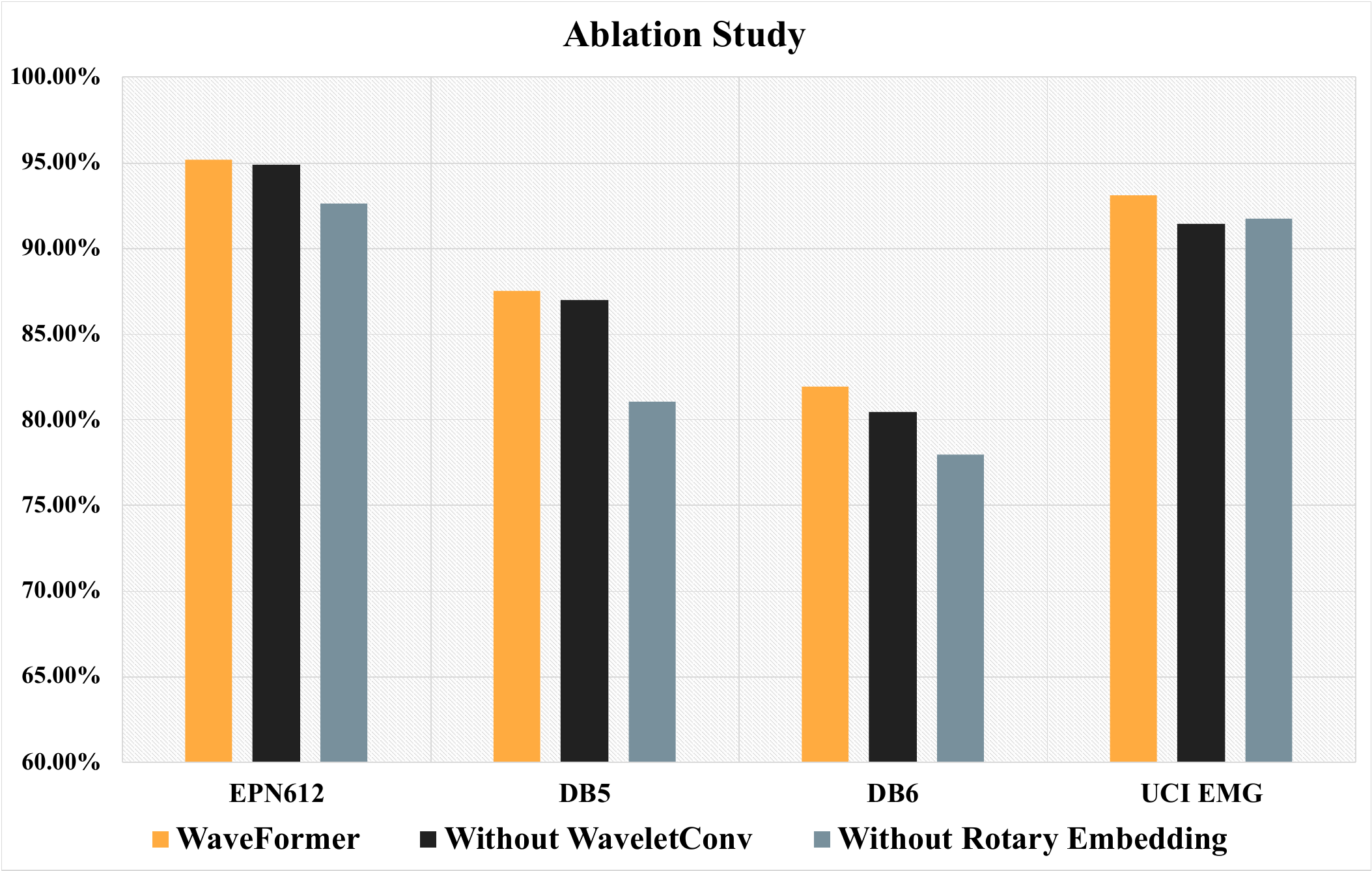}
  \caption{Ablation study comparing the full WaveFormer model with variants removing the WaveletConv module or rotary embedding across four downstream sEMG datasets.}
  \label{fig:ablation}
\end{figure}

We conducted ablation experiments to evaluate two key components: the WaveletConv module and rotary positional embedding (RoPE). As shown in Fig.~\ref{fig:ablation}, removing WaveletConv causes modest accuracy drops (0.33\%-1.65\%), while removing RoPE leads to more substantial degradation, particularly in DB5 where accuracy falls from 87.53\% to 81.08\% (-6.45\%). These results demonstrate that WaveletConv effectively captures multi-scale temporal features, while RoPE is crucial for positional awareness in complex multi-class scenarios. Together, these components enable the synergy between wavelet-domain processing and enhanced positional encoding for robust sEMG gesture recognition.

\subsection{Deployment Efficiency and Model Quantization}

We evaluated the computational efficiency of WaveFormer through comprehensive benchmarking on CPU hardware. The model was first converted to the ONNX format and then quantized to INT8 precision for optimization of deployment. The testing was carried out on an Intel Core i7-11800H processor using Python with ONNX Runtime. Following a 10-iteration warm-up, we measured performance over 200 inference iterations.

INT8 quantization achieves significant speedup with minimal accuracy loss. The quantized model delivers an average latency 6.75\,ms per sample, which is a 34\% reduction from the FP32 baseline (10.23\,ms). The throughput increases from 97.8 to 148.1 queries per second, representing an improvement 51\%. Memory efficiency also improves substantially, with CPU memory usage dropping from 84.9\,MB to 65.2\,MB. These results demonstrate that WaveFormer, with only 3.10 M parameters, achieves practical deployment efficiency on standard CPU hardware for real‑time sEMG processing systems.

\section{CONCLUSIONS}

This work demonstrates that WaveFormer achieves state-of-the-art sEMG gesture recognition with only 3.1 million parameters, proving that careful architectural design can eliminate the traditional trade-off between model size and accuracy. The integration of learnable wavelet transforms and RoPE attention enables real-time deployment (6.75ms latency) suitable for practical prosthetic control and rehabilitation applications. Future work will explore the adaptation and extension of the cross-subjects to other biomedical signals, potentially establishing WaveFormer as a unified framework for efficient neural signal processing in resource-constrained clinical environments.

\section*{ETHICAL APPROVAL}
The sEMG datasets used in this study were collected following approved ethical protocols. Original data collection procedures involving human subjects were approved by the respective Institutional Review Boards of the institutions where the data was collected. All participants provided their informed consent and the experimental procedures followed the principles outlined in the Declaration of Helsinki. No additional human subject experiments were performed specifically for this study.

\addtolength{\textheight}{-12cm}  




\section*{ACKNOWLEDGMENT}
This research was supported by the EU Horizon Europe project IntelliMan (grant agreement 101070136), the PNRR MUR project ECS00000033 ECOSISTER, the ETH Zürich’s Future Computing Laboratory funded by a donation from Huawei Technologies, and also the EU Horizon Europe project HAL4SDV (grant agreement 101139789). In addition, we gratefully acknowledge the computational resources made available by Cineca and CSCS. In particular, this work utilised the GPU-based high-performance computing infrastructures provided by Cineca and CSCS, whose support and operational services were instrumental in enabling the large-scale simulations/data-analysis undertaken for this study. 

\bibliographystyle{IEEEtran}
\bibliography{references}

\begin{thebibliography}{10}
\providecommand{\url}[1]{#1}
\csname url@samestyle\endcsname
\providecommand{\newblock}{\relax}
\providecommand{\bibinfo}[2]{#2}
\providecommand{\BIBentrySTDinterwordspacing}{\spaceskip=0pt\relax}
\providecommand{\BIBentryALTinterwordstretchfactor}{4}
\providecommand{\BIBentryALTinterwordspacing}{\spaceskip=\fontdimen2\font plus
\BIBentryALTinterwordstretchfactor\fontdimen3\font minus \fontdimen4\font\relax}
\providecommand{\BIBforeignlanguage}[2]{{%
\expandafter\ifx\csname l@#1\endcsname\relax
\typeout{** WARNING: IEEEtran.bst: No hyphenation pattern has been}%
\typeout{** loaded for the language `#1'. Using the pattern for}%
\typeout{** the default language instead.}%
\else
\language=\csname l@#1\endcsname
\fi
#2}}
\providecommand{\BIBdecl}{\relax}
\BIBdecl

\bibitem{scheme2011selective}
E.~J. Scheme, K.~B. Englehart, and B.~S. Hudgins, ``Selective classification for improved robustness of myoelectric control under nonideal conditions,'' \emph{IEEE Transactions on Biomedical Engineering}, vol.~58, no.~6, pp. 1698--1705, 2011.

\bibitem{wu2022electrode}
L.~Wu, A.~Liu, X.~Zhang, X.~Chen, and X.~Chen, ``Electrode shift robust cnn for high-density myoelectric pattern recognition control,'' \emph{IEEE Transactions on Instrumentation and Measurement}, vol.~71, pp. 1--10, 2022.

\bibitem{samuel2019intelligent}
O.~W. Samuel, M.~G. Asogbon, Y.~Geng, A.~H. Al-Timemy, S.~Pirbhulal, N.~Ji, S.~Chen, P.~Fang, and G.~Li, ``Intelligent emg pattern recognition control method for upper-limb multifunctional prostheses: advances, current challenges, and future prospects,'' \emph{Ieee Access}, vol.~7, pp. 10\,150--10\,165, 2019.

\bibitem{fajardo2021emg}
J.~M. Fajardo, O.~Gomez, and F.~Prieto, ``Emg hand gesture classification using handcrafted and deep features,'' \emph{Biomedical Signal Processing and Control}, vol.~63, p. 102210, 2021.

\bibitem{zanghieri2020temponet}
M.~Zanghieri, S.~Benatti, A.~Burrello, V.~Kartsch, F.~Conti, and L.~Benini, ``Robust real-time embedded emg recognition framework using temporal convolutional networks on a multicore iot processor,'' \emph{IEEE Transactions on Biomedical Circuits and Systems}, vol.~14, no.~2, pp. 244--256, 2020.

\bibitem{bakirciouglu2020classification}
K.~Bak{\i}rc{\i}o{\u{g}}lu and N.~{\"O}zkurt, ``Classification of emg signals using convolution neural network,'' \emph{International Journal of Applied Mathematics Electronics and Computers}, vol.~8, no.~4, pp. 115--119, 2020.

\bibitem{tsuji2000pattern}
T.~Tsuji, O.~Fukuda, M.~Kaneko, and K.~Ito, ``Pattern classification of time-series emg signals using neural networks,'' \emph{International Journal of Adaptive Control and Signal Processing}, vol.~14, no.~8, pp. 829--848, 2000.

\bibitem{goswami2024moment}
M.~Goswami, K.~Szafer, A.~Choudhry, Y.~Cai, S.~Li, and A.~Dubrawski, ``Moment: A family of open time-series foundation models,'' \emph{arXiv preprint arXiv:2402.03885}, 2024.

\bibitem{turgut2025generalisabletimeseriesunderstanding}
\BIBentryALTinterwordspacing
Özgün Turgut, P.~Müller, M.~J. Menten, and D.~Rueckert, ``Towards generalisable time series understanding across domains,'' 2025. [Online]. Available: \url{https://arxiv.org/abs/2410.07299}
\BIBentrySTDinterwordspacing

\bibitem{wang2014analysis}
G.~Wang, Y.~Zhang, and J.~Wang, ``The analysis of surface emg signals with the wavelet-based correlation dimension method,'' \emph{Computational and mathematical methods in medicine}, vol. 2014, no.~1, p. 284308, 2014.

\bibitem{zhang2019wavelet}
D.~Zhang and D.~Zhang, ``Wavelet transform,'' \emph{Fundamentals of image data mining: Analysis, Features, Classification and Retrieval}, pp. 35--44, 2019.

\bibitem{finder2024wavelet}
S.~E. Finder, R.~Amoyal, E.~Treister, and O.~Freifeld, ``Wavelet convolutions for large receptive fields,'' in \emph{European Conference on Computer Vision}.\hskip 1em plus 0.5em minus 0.4em\relax Springer, 2024, pp. 363--380.

\bibitem{su2024roformer}
J.~Su, M.~Ahmed, Y.~Lu, S.~Pan, W.~Bo, and Y.~Liu, ``Roformer: Enhanced transformer with rotary position embedding,'' \emph{Neurocomputing}, vol. 568, p. 127063, 2024.

\bibitem{heo2024rotary}
B.~Heo, S.~Park, D.~Han, and S.~Yun, ``Rotary position embedding for vision transformer,'' in \emph{European Conference on Computer Vision}.\hskip 1em plus 0.5em minus 0.4em\relax Springer, 2024, pp. 289--305.

\bibitem{frusque2024robust}
G.~Frusque and O.~Fink, ``Robust time series denoising with learnable wavelet packet transform,'' \emph{Advanced Engineering Informatics}, vol.~62, p. 102669, 2024.

\bibitem{gao2024efficient}
X.~Gao, T.~Qiu, X.~Zhang, H.~Bai, K.~Liu, X.~Huang, H.~Wei, G.~Zhang, and H.~Liu, ``Efficient multi-scale network with learnable discrete wavelet transform for blind motion deblurring,'' in \emph{Proceedings of the IEEE/CVF Conference on Computer Vision and Pattern Recognition}, 2024, pp. 2733--2742.

\bibitem{vonesch2007generalized}
C.~Vonesch, T.~Blu, and M.~Unser, ``Generalized daubechies wavelet families,'' \emph{IEEE transactions on signal processing}, vol.~55, no.~9, pp. 4415--4429, 2007.

\bibitem{boyer2023reducing}
M.~Boyer, L.~Bouyer, J.-S. Roy, and A.~Campeau-Lecours, ``Reducing noise, artifacts and interference in single-channel emg signals: A review,'' \emph{Sensors}, vol.~23, no.~6, p. 2927, 2023.

\bibitem{eddy2024big}
E.~Eddy, E.~Campbell, S.~Bateman, and E.~Scheme, ``Big data in myoelectric control: large multi-user models enable robust zero-shot emg-based discrete gesture recognition,'' \emph{Frontiers in Bioengineering and Biotechnology}, vol.~12, p. 1463377, 2024.

\bibitem{pizzolato2017comparison}
S.~Pizzolato, L.~Tagliapietra, M.~Cognolato, M.~Reggiani, H.~M{\"u}ller, and M.~Atzori, ``Comparison of six electromyography acquisition setups on hand movement classification tasks,'' \emph{PloS one}, vol.~12, no.~10, p. e0186132, 2017.

\bibitem{palermo2017repeatability}
F.~Palermo, M.~Cognolato, A.~Gijsberts, H.~M{\"u}ller, B.~Caputo, and M.~Atzori, ``Repeatability of grasp recognition for robotic hand prosthesis control based on semg data,'' in \emph{2017 International Conference on Rehabilitation Robotics (ICORR)}.\hskip 1em plus 0.5em minus 0.4em\relax IEEE, 2017, pp. 1154--1159.

\bibitem{karapinar2021machine}
Z.~K. Senturk, M.~S. Bakay \emph{et~al.}, ``Machine learning-based hand gesture recognition via emg data,'' \emph{ADCAIJ: Advances in Distributed Computing and Artificial Intelligence Journal}, vol.~10, no.~2, pp. 123--136, 2021.

\bibitem{li2024high}
H.~Li, Y.~Li, J.~Luo, X.~Jiao, J.~Liu, L.~Zhou, L.~Chang, and J.~Zhou, ``A high accuracy and real-time semg-based hand gesture classifier using lda-based template matching with adaptive majority vote and online data augmentation,'' \emph{IEEE Sensors Journal}, 2024.

\bibitem{burrello2022bioformers}
A.~Burrello, F.~B. Morghet, M.~Scherer, S.~Benatti, L.~Benini, E.~Macii, M.~Poncino, and D.~J. Pagliari, ``Bioformers: Embedding transformers for ultra-low power semg-based gesture recognition,'' in \emph{2022 Design, Automation \& Test in Europe Conference \& Exhibition (DATE)}.\hskip 1em plus 0.5em minus 0.4em\relax IEEE, 2022, pp. 1443--1448.

\end{thebibliography}
\end{document}